\documentclass{amia}
\usepackage{graphicx}
\usepackage[labelfont=bf]{caption}
\usepackage[superscript,nomove]{cite}
\usepackage{color}
\usepackage{url}
\usepackage{booktabs}
\usepackage{multirow}
\usepackage[super]{nth}

\newlength{\Tstrut}
\newcommand\Hstrut{\rule{3em}{0pt}}
\newcommand{\pattern}[1]{\texttt{#1}}

\begin{document}

\title{NegBio: a high-performance tool for negation and uncertainty detection in radiology reports}

\author{Yifan Peng, Ph.D.$^{1}$, Xiaosong Wang, Ph.D.$^{2}$, Le Lu, Ph.D.$^{2}$, Mohammadhadi Bagheri, M.D.$^{2}$, Ronald Summers, M.D., Ph.D.$^{2}$, Zhiyong Lu, Ph.D.$^{1}$}

\institutes{
    $^1$National Center for Biotechnology Information, National Library of Medicine, National Institutes of Health, Bethesda, MD, USA; $^2$Department of Radiology and Imaging Sciences, Clinical Center, National Institutes of Health, Bethesda, MD, USA\\
}

\maketitle

\noindent{\bf Abstract}

\textit{Negative and uncertain medical findings are frequent in radiology reports, but discriminating them from positive findings remains challenging for information extraction. Here, we propose a new algorithm, NegBio, to detect negative and uncertain findings in radiology reports. Unlike previous rule-based methods, NegBio utilizes patterns on universal dependencies to identify the scope of triggers that are indicative of negation or uncertainty. We evaluated NegBio on four datasets, including two public benchmarking corpora of radiology reports, a new radiology corpus that we annotated for this work, and a public corpus of general clinical texts. Evaluation on these datasets demonstrates that NegBio is highly accurate for detecting negative and uncertain findings and compares favorably to a widely-used state-of-the-art system NegEx (an average of 9.5\% improvement in precision and 5.1\% in F1-score).}

Availability: \url{https://github.com/ncbi-nlp/NegBio}

\section*{Introduction}

In radiology, findings are observations regarding each area of the body examined in the imaging study and their mentions in radiology reports can be positive, negative or uncertain. In this paper, we call a finding negative if it is negated, and uncertain if in an equivocal or hypothetical statement. For example, ``pneumothorax'' is \textit{negative} in ``no evidence of pneumothorax'' and is \textit{uncertain} in ``suspicious pneumothorax''. 

Negative and uncertain findings are frequent in radiology reports~\cite{chapman2001evaluation}. Since they may indicate the absence of findings mentioned within the radiology report, identifying them is as important as identifying positive findings. Otherwise, information extraction algorithms that do not distinguish negative and uncertain findings from positive ones may return many irrelevant results. Even though many natural language processing applications have been developed in recent years that successfully extract findings mentioned in medical reports, discriminating between positive, negative, and uncertain findings remains challenging~\cite{wu2014negations,gkotsis2016dont,morante2010descriptive,wu2011evaluation}.
 
Previous efforts in this area include both rule-based and machine-learning approaches. Rule-based systems rely on negation keywords and rules to determine the negation~\cite{friedman1999representing}. NegEx is a widely used algorithm that utilizes regular expressions~\cite{chapman2001simple,harkema2009context}. However, regular expressions rely solely on surface text, and thus are limited when attempting to capture complex syntactic constructions such as long noun phrases. In its early version, NegEx limited the scope by hard-coded word windows size. For example, NegEx cannot detect negative ``effusion'' in ``clear of focal airspace disease, pneumothorax, or pleural effusion'' because ``effusion'' is beyond the scope of ``clear'' (5 words). In its later versions, the algorithm (ConText~\cite{chapman2011document}) extended scope to the end of the sentence (or allow the user to set a window size).  In this work, we use the NegEx enhanced version via MetaMap\cite{aronson2010overview}. 

In addition to regular expressions, there were proposals to use parse trees or dependency graph to capture long distance information between negation keywords and the target. However, none defined patterns directly on the syntactic structures to take the advantage of linguistic knowledge~\cite{mutalik2001use,sohn2012dependency,mehrabi2015deepen}. For example, Sohn~et~al, 2012) used regular expressions on the dependency path~\cite{sohn2012dependency} and (Mehrabi~et~al, 2015) used dependency patterns as a post-processing step after NegEx to remove false positives of negative findings~\cite{mehrabi2015deepen}. Moreover, none of these dependency graph-based methods is made publicly available. Finally, machine learning offers another approach to extract negations~\cite{wu2014negations,huang2007novel,clark2011mitre,goryachev2006implementation}. These approaches need manually annotated in-domain data to ensure their performance. Unfortunately, such data are generally not publicly available~\cite{ogren2008constructing,uzuner20112010,suominen2013overview,albright2013towards}. Furthermore, machine learning based approaches often suffer in generalizability – the ability to perform well on text previously unseen. 

In this work, we propose NegBio, a new and open-source rule-based tool for negation and uncertain detection in radiology reports. Unlike previous methods, NegBio utilizes universal dependencies for pattern definition and subgraph matching for graph traversal search so that the scope for negation/uncertainty is not limited to fixed word distance~\cite{de2014universal,liu2013approximate}. In addition to negation, NegBio also detects uncertainty, a useful feature that is not well studied before. 

For evaluating NegBio, we first assessed its ability to improve the correct extraction of positively asserted medical findings, which is a practical task where negation detection is often required. That is, NegBio is applied to remove negated and uncertain findings in an end-to-end information extraction system, which takes raw clinical text as input and aims to extract only positively asserted findings in its output. By doing so, we expect to see improvements in precision for the whole system. In the meantime, we compared NegBio with the widely-used NegEx system. 

For ensuring NegBio is a robust and generalizable approach, two data sets were used for this purpose. One is a public benchmarking dataset, OpenI~\cite{demner2015preparing}. The other is a newly created corpus, ChestX-ray, which includes 900 radiology with 14 informative yet generic types of medical findings. In both datasets, only positive findings are annotated. 

Furthermore, we also evaluated NegBio on its performance to detect negations in two additional corpora (BioScope~\cite{vincze2010speculation} and PK\footnote{\url{https://storage.googleapis.com/google-code-archive-downloads/v2/code.google.com/negex/negex.python.zip}}) where negated expressions were fully annotated. On BioScope, we followed the lead of (Demner-Fushman~et~al, 2017) in our evaluation by using MetaMap to annotate the negative findings and treat them as ground truth~\cite{demner-fushman2017metamap}. Also note that unlike the radiology reports in the other three corpora, the PK corpus consists of general clinical texts.

\section*{Methods}

\begin{figure}[h!]
\centering
\includegraphics[width=.9\textwidth,trim={1cm 13cm 5cm .5cm},clip]{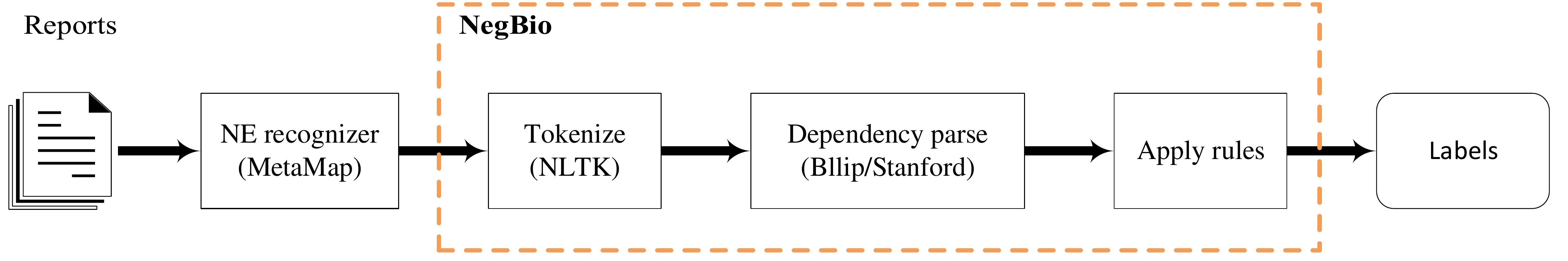}
\caption{An overall pipeline of NegBio.}
\label{fig:pipeline}
\end{figure}

NegBio tasks as inputting a sentence with pre-tagged mentions of medical findings, and checks whether a specific finding is negative or uncertain. Figure~\ref{fig:pipeline} shows the overall pipeline of NegBio. In the case of using MetaMap alone, the system will skip the NegBio part and produce the labels directly. Detailed steps are described in the following sub-sections.\\

\subsection*{Medical findings recognition}

Our approach labeled the reports in two passes. We first detected all the findings and their corresponding UMLS\textsuperscript{\textcopyright{}} concepts using MetaMap~\cite{aronson2010overview}. Here we only focused on 14 common disease finding types as described in Table~\ref{tab:detailed corpora}. The 14 finding types are most common in our institute, which are selected by radiologists from a clinical perspective. The next steps involved applying NegBio to all identified findings and subsequently ruling out those that are negative and uncertain.

\subsection*{Universal dependency graph construction}
\label{sec:udg}

\begin{figure}
\centering
\includegraphics[scale=1,trim={1cm 1cm 1.5cm 1.5cm},clip]{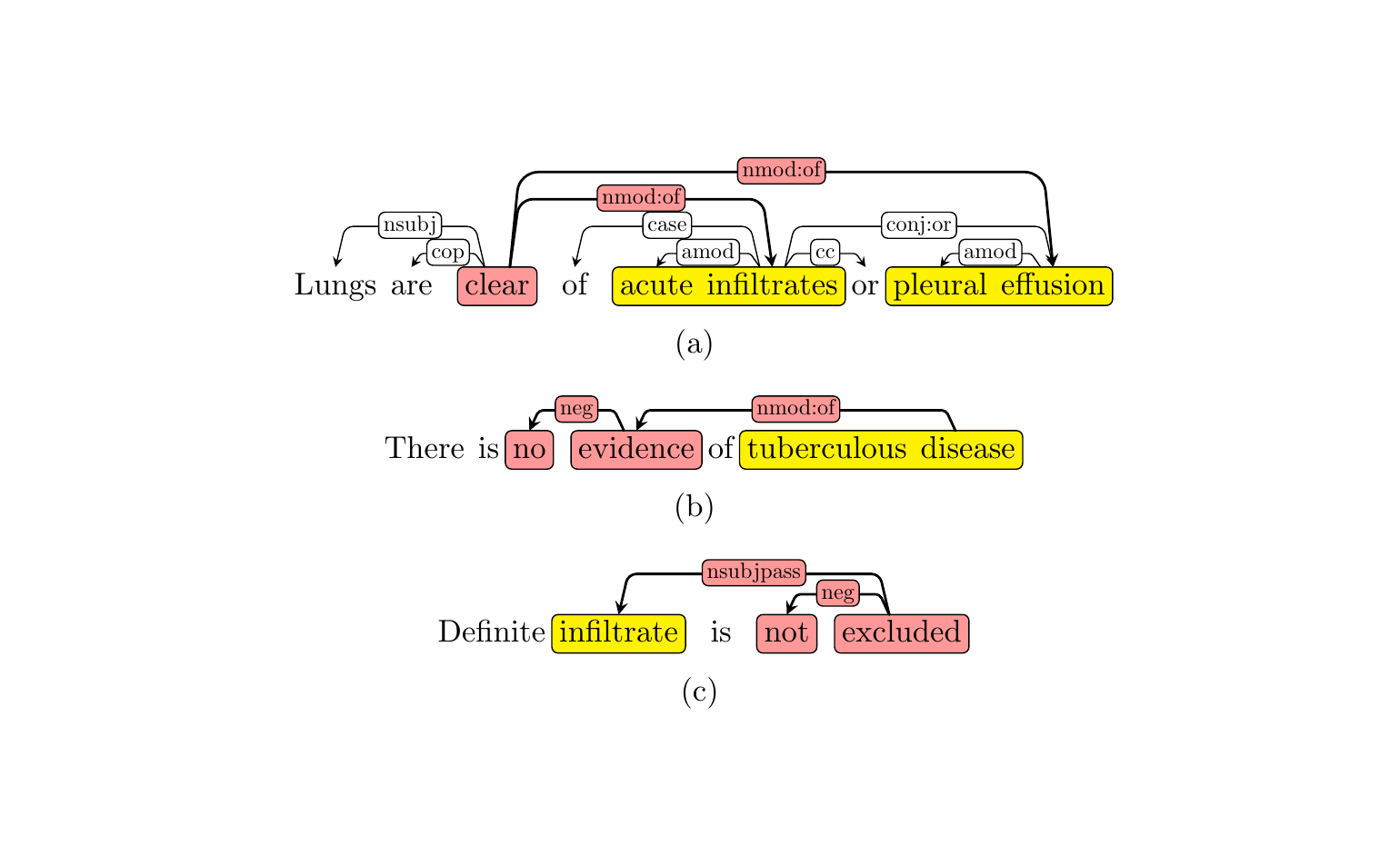}
\caption{The dependency graph of (a) ``Lungs are clear of acute infiltrates or pleural effusion'', (b) ``There is no evidence of tuberculous disease, and (c) ``Definite infiltrate is not excluded''.}
\label{fig:dg}
\end{figure}

In this work, we utilized the universal dependency graph to define patterns. It was designed to provide a simple description of the grammatical relationships in a sentence that can be easily understood by non-linguists and effectively used by downstream language understanding tasks. All universal dependencies information can be represented by a directed graph, called universal dependency graph (UDG). The vertices in a UDG are labeled with information such as the word, part-of-speech and the word lemma. The edges in a UDG represent typed dependencies from the governor to its dependent and are labeled with dependency type such as ``\pattern{nsubj}'' (nominal subject) or ``\pattern{conj}'' (conjunction). Figure~\ref{fig:dg}(a) shows a UDG of sentence ``Lungs are clear of acute infiltrates or pleural effusion.'' where ``Lung'' is the subject and ``acute infiltrates'' and ``pleural effusion'' are two coordinating findings.\\

To obtain the UDG of a sentence, we first split and tokenized each report into sentences using NLTK~\cite{bird2009natural}. Next, we parsed each sentence with the Bllip parser trained with the biomedical model~\cite{charniak2005coarse,mcclosky2009any}. The universal dependencies were then obtained by applying the Stanford dependencies converter on the parse tree with the \textit{CCProcessed} and \textit{Universal} option~\cite{de2014universal,demarneffe2015stanford}.  

\subsection*{Negation and uncertainty detection}

In NegBio, we defined rules on the UDG by utilizing the dependency label and direction information\footnote{Due to space restriction, we include the complete set of rules in the released source code.}. We searched the UDG from the head word of a finding mention (e.g., ``effusion'' in Figure~\ref{fig:dg}(a)). If the word node matches one of our pre-defined patterns, we treated it as negative/uncertain. For example, the finding ``pleural effusion'' in Figure~\ref{fig:dg}(a) is negated because it matches the rule ``\pattern{\{\} <nmod:of \{lemma:/clear/\}}''. The rule indicates that ``clear'' is the governor of ``effusion'' with a dependency ``\pattern{nmod:of}''. In NegBio, we use ``Semgrex'', a pattern language, for rapid development of dependency rules~\cite{chambers2007learning}. 

Rules in NegBio can be either a chain of dependencies or a sub-graph. Figure~\ref{fig:dg}(b) shows how a chain rule ``\pattern{\{\} <nmod:of (\{lemma:/evidence/\} <neg \{word:/no/\})}'' matches ``tuberculous disease'' as negative. Figure~\ref{fig:dg}(c) shows how a sub-graph rule ``\pattern{\{\} < (\{lemma:/exclude/\} >neg \{word:/not/\})}'' matches ``infiltrate''. We call the latter case sub-graph matching because ``not'' is attached to ``excluded'' and is the sibling of ``infiltrate'' in the UDG. 

Since our patterns are defined on the graph, the negation/uncertainty scope is thus not limited to word distance. Instead, it is based on syntactic context. Specifically, we converted each rule to a subgraph for matching nodes/edges in the dependency graph. Here, we applied sub-graph matching algorithm to search the patterns in the graph~\cite{liu2013approximate}. Therefore, the negation/uncertainty scope is all vertices covered in the subgraph. The computational complexity of the Bllip parser and subgraph-matching algorithm is $O(m^3)$ and $O(m^2k^m)$ respectively where $m$ is the length of an input sentence and $k$ is the vertex degree.

\section*{Results}

\subsection*{Evaluations on findings detection in an end-to-end system}

First, we evaluated NegBio by comparing the final extracted findings to the gold standard. Precision, recall, and F1-score were computed accordingly based on the number of true positives, false positives, and false negatives.
 
We used the following dataset to evaluate NegBio negation/uncertainty detection (\nth{1} and \nth{2} rows in Table~\ref{tab:corpora}).\\
\begin{table}[h]
\centering
\caption{Descriptions of OpenI, ChestX-ray, BioScope, and PK.\label{tab:corpora}}
\begin{tabular}{lrrr}
\toprule
Dataset    & Reports  & Positives & Negatives\\
\midrule
OpenI & 3,851 & 1,354 & --\\
ChestX-ray & 900 & 2,131 & --\\
BioScope test set & 977 & -- & 466\\
PK & 116 & -- & 491\\
\bottomrule
\end{tabular}
\end{table}

\textbf{OpenI} is a publicly available radiology dataset~\cite{demner2015preparing}. Using the OpenI API\footnote{\url{https://openi.nlm.nih.gov/retrieve.php?it=x&coll=iu}}, we retrieved 3,851 unique radiology reports where each OpenI report was annotated with key concepts including body parts (e.g., ``lung''), findings (e.g., ``pneumothorax'') and diagnoses (e.g., ``tuberculosis''). Then, the radiologist (Bagheri M) manually checked the annotations in OpenI and explicitly distinguished between a body part, a finding and a diagnosis. The findings were then organized into fine-grained categories for two reasons. First, each category should have enough examples for the evaluation. Second, such categories should contain enough details to facilitate correlation of findings with the diagnosis. As a result, we obtained 14 domain-important yet generic types of medical findings, enabling computational inference from symptoms to a disease in the future. The final dataset is shown in Table~\ref{tab:corpora} and Table~\ref{tab:detailed corpora}.\\
\begin{table}[h]
\centering
\caption{Number of findings in OpenI and ChestX-ray.\label{tab:detailed corpora}}
\begin{tabular}{lrr}
\toprule
Finding    & OpenI  & ChestX-ray\\
\midrule
Atelectasis & 315 & 311\\
Cardiomegaly & 345 & 202\\
Consolidation & 30 & 79\\
Edema & 42 & 43\\
Effusion & 155 & 381\\
Emphysema & 103 & 54\\
Fibrosis & 23 & 15\\
Hernia & 46 & 2\\
Infiltration & 60 & 383\\
Mass & 15 & 114\\
Nodule & 106 & 154\\
Pleural Thickening & 52 & 52\\
Pneumonia & 40 & 62\\
Pneumothorax & 22 & 279\\
\rule{2em}{0pt}\textit{Total} & 1,354 & 2,131\\
\bottomrule
\end{tabular}
\end{table}

\textbf{ChestX-ray} is a newly constructed gold-standard dataset to assess the robustness of NegBio~\cite{wang2017chestx}. We randomly selected 900 reports from a larger radiology dataset collected from a national hospital and asked two annotators to mark the above 14 types of findings. A trial set of 30 reports was first used to help us better understand the annotation task. Then, each report was independently annotated by two experts. In this paper, we used the inter-rater agreement (IRA) to measure the level of agreement between two experts~\cite{mchugh2012interrater}. The Cohen's kappa is 84.3\%.

In the experiments, we considered three scenarios: using (1)~MetaMap only, (2)~MetaMap with the baseline algorithm NegEx, and (3)~MetaMap with NegBio. In this paper, we used 80\% of the OpenI to design the patterns, and used the remaining 20\% for testing. We did not further tune the NegBio patterns for ChestX-ray, thus the full dataset was used for testing. Please also note that ``negative'' and ``uncertain'' cases are not annotated on the document level in both OpenI and ChestX-ray. 

Table~\ref{tab:evaluation positive} shows the results on OpenI and ChestX-ray corpora, as measured by precision~(P), recall~(R), and F1-score~(F). The \nth{1} and \nth{2} rows demonstrate that negation/uncertainty detection dramatically improves the precision (from 13.8\% to 77.2\%), even with a baseline approach. As a result, the F1-score also increases significantly (from 23.8\% to 80.7\%). This observation proves the usefulness of negation detection in the task of information extraction. The 2nd and 3rd rows compare NegEx with our method NegBio. Overall, NegBio achieved a higher precision of 89.8\%, recall of 85.0\%, and F1-score of 87.3\% on OpenI.\\
\begin{table}[h]
\centering
\caption{Evaluation results on OpenI, ChestX-ray using (1) MetaMap, (2) MetaMap and NegEx, and (3) MetaMap and NegBio. Performance is measured by precision (P), recall (R), and F1-score (F) on positive findings.\label{tab:evaluation positive}}
\begin{tabular}{p{8em}rrrrrr}
\toprule
\multirow{2}{*}{Method} & \multicolumn{3}{c}{OpenI} & \multicolumn{3}{c}{ChestX-ray}\\
\cmidrule(lr){2-4}\cmidrule(l){5-7}
& \Hstrut{}P & \Hstrut{}R & \Hstrut{}F & \Hstrut{}P & \Hstrut{}R & \Hstrut{}F\\
\midrule
MetaMap & 13.8 & \textbf{85.7} & 23.8 & 72.3 & \textbf{95.7} & 82.4\\
MetaMap+NegEx & 77.2 & 84.6 & 80.7 & 82.8 & 95.5 & 88.7\\
MetaMap+NegBio & \textbf{89.8} & 85.0 & \textbf{87.3} & \textbf{94.4} & 94.4 & \textbf{94.4}\\
\bottomrule
\end{tabular}
\end{table} 

To test the generalizability of NegBio, we repeated the experiments on the second dataset ChestX-ray. We observed that on ChestX-ray, the overall precision with NegBio was substantially higher (11.6\% improvement) than that of NegEx with comparable recall (94.4\%) and overall higher F1-score (94.4\%). 

\subsection*{Experiments on negation detection}

On these two datasets, we evaluated NegBio by comparing the ``negations'' recognition results to the gold standard. In other words, the extracted results are considered a true positive if they are annotated as negative in the document. We used the following dataset to evaluate NegBio negation/uncertainty detection (\nth{3} and \nth{4} rows in Table~\ref{tab:corpora}).

\textbf{BioScope} consists of medical and biological texts annotated for negation, speculation and their linguistic scope~\cite{vincze2010speculation}. Here, we considered only negation annotations in BioScope for our purpose. Hence, the test set of medical free-texts consists of 977 radiology reports with 466 negative scopes. To set the ground truth for negations, we followed the lead of (Demner-Fushman~et~al, 2017) in our evaluation~\cite{demner-fushman2017metamap}. We used the MetaMap to annotate the findings in the negative scopes and treat them as ground truth. As a result, we obtained 233 findings within the 466 annotated negative scopes.

\textbf{PK} (prepared by Peter Kang) consists of 116 documents with 1,885 ``affirmed'' and 491 ``negated'' phrases. Different from OpenI, ChestX-ray, and BioScope that are all radiological reports, PK consists of general clinical text thus is suitable to test the generalizability of NegBio on other types of clinical texts.

Table~\ref{tab:evaluation negative} shows the results on BioScope and PK respectively. On BioScope, we observed that, the overall performance of NegBio was higher than that of NegEx with a substantial 25.5\% increase in precision and 13.6\% increase in F1-score. On PK, NegBio also achieved slightly higher precision (2.7\%) and F1-scores (0.2\%).\\
\begin{table}[h]
\centering
\caption{Evaluation results on BioScope and PK using NegEx and NegBio. Performance is measured by precision~(P), recall~(R), and F1-score~(F) on negations.\label{tab:evaluation negative}}
\begin{tabular}{p{8em}rrrrrr}
\toprule
\multirow{2}{*}{Method} & \multicolumn{3}{c}{BioScope} & \multicolumn{3}{c}{PK}\\
\cmidrule(lr){2-4}\cmidrule(l){5-7}
& \Hstrut{}P & \Hstrut{}R & \Hstrut{}F & \Hstrut{}P & \Hstrut{}R & \Hstrut{}F\\
\midrule
NegEx & 70.6 & \textbf{98.7} & 82.3 & 95.1 & \textbf{91.2} & 93.1\\
NegBio & \textbf{96.1} & 95.7 & \textbf{95.9} & \textbf{98.4} & 88.6 & \textbf{93.3}\\
\bottomrule
\end{tabular}
\end{table} 

\section*{Discussion}

Overall, NegBio achieved a significant improvement on all datasets over the popular method NegEx. This indicates that the use of negation and uncertainty detection on the syntactic level successfully removes false positive cases of ``positive'' findings. 

In general, NegBio leverages syntactic structures in the rules. Hence, its rules are expected to be not only stricter than the regular expressions but also more generalizable to match more text variations. In the negation detection task, NegBio achieved higher precision because the patterns are stricter. For example, the ``difficult to keep focused'' is positive in ``His review of systems is limited by the fact that he is not terribly cooperative and he is difficult to keep focused''. The regular expression ``\pattern{not .*}'' used in NegEx over-extends the negation scope of ``not'' to the end of the sentence. Therefore, NegEx incorrectly detects ``difficult to keep focused'' as a negative. On the other hand, NegBio can detect that the negation scope of ``not'' is ``terribly cooperative'' according to the conjunction syntactic structure of this sentence. Therefore, NegBio correctly detects ``difficult to keep focused'' as a positive. As for the recall, NegBio did not achieve higher recall because both datasets are relatively small and contain limited text variation. 

In the positive findings detection tasks, the recalls of NegBio are comparable to NegEx because we count positive findings on the document level. In other words, even if the NegBio patterns miss one negation in one sentence, it might detect others in the same document. More interestingly, NegBio achieved higher precision for two main reasons. One is due to the uncertainty detection. We further assessed the effects of uncertain finding detection rules. When disabled in NegBio, the overall performance dropped 7.4\% and 2.5\% relatively in F-score on OpenI and ChestX-ray datasets respectively. The results demonstrate that uncertain finding detection is important in this task. The second reason is due to the text variations. Comparing the size of our four datasets, OpenI and ChestX-ray are much larger, in terms of unique mentions, than BioScope and PK. Therefore, they contain more text variations.
 
Furthermore, we analyzed the errors of our method and categorized them into three major types. The first type is related to Named Entity Recognition accuracy where some findings are difficult to be recognized correctly by MetaMap. For example, 30\% of ``Nodule'' were not correctly recognized on the OpenI corpus. 

The second type of errors is due to parsing, on which our patterns rely on input. In X-ray reports, instead of using a full sentence, radiologists tend to use long noun phrases without verbs to describe the findings (e.g., ``No definite pleural effusion seen, no typical findings of pulmonary edema.''). The Bllip parser could not parse these noun phrases accurately. 
 
Finally, our rules missed double negation (e.g., ``Findings cannot exclude increasing pleural effusions.''). The double negation cancels one another and indicates a positive finding. Our method currently lacks rules to recognize double negatives and thus generates more false negatives. While there are studies discussing this topic by providing limited double-negation triggers~\cite{nassif2009information,chapman2013extending}, it is not yet known if they are generalizable to complex sentences and applicable on the dependency structure. We therefore regard double-negation as an open challenge that warrants further investigation.
 
We also noticed that MetaMap had a much lower performance on OpenI than its results on ChestX-ray without negation detection. This is largely due to its low precisions on ``Pneumothorax'' and ``Consolidation'' in the OpenI dataset. Since only positive findings are annotated in OpenI and ChestX-ray, we are unable to evaluate and report all the ``negation'' and ``uncertain'' cases in these two corpora. Fully annotating all the negations and uncertainties in these two corpora remains as future work for us.  

\section*{Conclusion}

In this paper, we propose an algorithm, NegBio, to determine negative and uncertain findings in radiology reports. This information is also useful for improving the precision of information extraction from radiology reports. We evaluated NegBio on two publicly available corpora and a newly constructed corpus. We showed that NegBio achieved a significant improvement on all datasets over the state of the art. By making NegBio an open source tool, we believe it can contribute to the research and development in healthcare informatics community for real-world applications. In the future, we plan to explore its applicability in clinical texts beyond radiology reports.

\section*{Acknowledgements}
This work was supported by the Intramural Research Programs of the National Institutes of Health, National Library of Medicine and Clinical Center. We are also grateful to the authors of NegEx, DNorm, and MetaMap for making their software tools publicly available, Drs. Dina Demner-Fushman and Willie J Rogers for the helpful discussion, and the NIH Fellows Editorial Board for their editorial comments.

\makeatletter
\renewcommand{\@biblabel}[1]{\hfill #1.}
\makeatother

\bibliography{reference}
\bibliographystyle{unsrt}
%
%
%
%

\end{document}